\pdfoutput=1

\documentclass[11pt]{article}

\usepackage[final]{ACL2023}
\usepackage{comment}
\usepackage{times}
\usepackage{latexsym}
\usepackage{inconsolata}
\usepackage{array} 
\usepackage{ragged2e}
\usepackage{caption} 
\usepackage{booktabs}
\usepackage[utf8]{inputenc}
\usepackage{amssymb}
\usepackage{graphicx}
\usepackage{amsmath}
\usepackage{todonotes}
\usepackage{tcolorbox}

\usepackage{CJKutf8} 

\usepackage{subfigure}
\usepackage[T1]{fontenc}

\usepackage[utf8]{inputenc}

\usepackage{microtype}

\usepackage{inconsolata}

\title{What Language(s) Does Aya-23 Think In? How Multilinguality Affects Internal Language Representations}

\author{
Katharina Trinley\textsuperscript{1,}\thanks{*Equal contribution}  ,
Toshiki Nakai\textsuperscript{1, }\textcolor{darkblue}{$^*$},
Tatiana Anikina\textsuperscript{2}, 
Tanja Baeumel\textsuperscript{2} \\
\textsuperscript{1}Saarland University \\
\textsuperscript{2}German Research Center for Artificial Intelligence (DFKI) \\
\vspace{0.5em}
\texttt{\{katr00001, tona00002\}@stud.uni-saarland.de} \\
}

\begin{document}
\maketitle

\begin{abstract}
Large language models (LLMs) excel at multilingual tasks, yet their internal language processing remains poorly understood. We analyze how Aya-23-8B, a decoder-only LLM trained on balanced multilingual data, handles code-mixed, cloze, and translation tasks compared to predominantly monolingual models like Llama 3 and Chinese-LLaMA-2. Using logit lens and neuron specialization analyses, we find: (1) Aya-23 activates typologically related language representations during translation, unlike English-centric models that rely on a single pivot language; (2) code-mixed neuron activation patterns vary with mixing rates and are shaped more by the base language than the mixed-in one; and (3) Aya-23's language-specific neurons for code-mixed inputs concentrate in final layers, diverging from prior findings on decoder-only models. Neuron overlap analysis further shows that script similarity and typological relations impact processing across model types. These findings reveal how multilingual training shapes LLM internals and inform future cross-lingual transfer research.
\end{abstract}

\section{Introduction}
Large language models (LLMs) excel in multilingual tasks \cite{srivastava2022beyond, bang2023multitask, gurgurov2025smallmodelsbigimpact}, but their internal handling of multiple languages remains underexplored \cite{kaddour2023challenges}. While methods like logit lens \cite{wendler2024llamas, schut2025multilingual} and neuron specialization \cite{tang2024language, kojima2024multilingual, tan2024neuron} have been applied, prior work mainly targets English-centered models on monolingual tasks (e.g., cloze or repetition tasks), rather than balanced multilingual architectures and their processing of code-mixed texts.

Multilingual models often default to English during intermediate processing, as described by the Multilingual Workflow (MWork) hypothesis \cite{zhao2024large}, which suggests LLMs convert non-English inputs into English internally before generating outputs. Supporting this, studies on reasoning language models (RLMs) \citep{wang2025language} find reliance on internal “pivot” languages or scripts, even with other input languages. However, it remains unclear if this preference is unique to RLMs or a general pattern in all multilingual LLMs. Therefore, we ask:

\newcommand{\researchq}[2]{%
\begin{tcolorbox}[
    colback=gray!8, 
    colframe=gray!50, 
    boxrule=0.4pt, 
    sharp corners,
    left=3pt, 
    right=3pt, 
    top=2pt, 
    bottom=2pt,
    fontupper=\small
]
\textbf{#1:} #2
\end{tcolorbox}
}

\researchq{H1}{How do balanced multilingual models process translation tasks -- do they activate multiple languages simultaneously, unlike English-centric models that rely on a single pivot language?}

Neuron-level analyses have identified language-specific patterns \cite{kojima2024multilingual, tang2024language}, but these studies predominantly examine English-based models, leaving open whether multilingual training leads to fundamentally different internal processing mechanisms. While LLMs' language capabilities are tied to specific neuron subsets, particularly in early and late layers \cite{kojima2024multilingual, tang2024language}, these patterns may not apply to models trained on diverse multilingual data \cite{zhong2024beyond, schut2025multilingual}. We thus investigate the following hypotheses:

\researchq{H2}{What patterns of neuron sharing of language specific neurons emerge in balanced multilingual models, and do these align more strongly with language similarity compared to predominantly monolingual models?}

\researchq{H3}{Where do language-specific neurons concentrate in multilingual architectures -- do they cluster predominantly in final layers, contrary to prior findings showing distribution across early and late layers in decoder-only models?}

In real-world contexts, speakers often mix languages within a single utterance, requiring models to dynamically switch between language-specific representations. Code mixing (CM) provides a valuable lens for studying multilingual processing in language models \cite{xie2025enhancing}, and while multilingual LLMs perform well on some tasks, they still struggle with code-switched text \cite{gundapu2020word}. The development of more balanced multilingual models, such as Aya-23 \cite{aryabumi2024aya}, offers an opportunity to examine how different training approaches affect internal language representations, especially when handling the linguistic complexity of code-mixed inputs. Thus, we ask:






\researchq{H4}{How does the processing of code-mixed inputs vary based on language pair characteristics and models?}

To address these questions, we perform a neuron-level comparison of a balanced multilingual model (Aya-23-8B), a predominantly English-trained model (Llama 3.1-8B), and a language-specialized model (Chinese-LLaMA-2-7B). Specifically, we:

\paragraph{I. Analyze internal language representations} across 13-language translation tasks using logit lens to test \textbf{H1}, checking whether Aya-23 activates multiple languages simultaneously, unlike English-pivot processing in mostly monolingual models.

\paragraph{II. Create a controlled code-mixed dataset} with varying mixing ratios across 10 typologically diverse pairs (\{fr, zh\} $\times$ \{en, es, it, ja, ko\}) and use neuron specialization (activation frequency \cite{tan2024neuron}) to investigate \textbf{H2} and \textbf{H4}, exploring how script similarity and language relationships affect neuron sharing across models.

\paragraph{III. Examine layer-wise distribution of language-specific neurons} via activation strength \cite{kojima2024multilingual} to test \textbf{H3}, determining whether balanced multilingual training concentrates language-specific neurons mainly in final layers, contrasting prior findings of early-and-late layer distributions in decoder-only models.


\section{Methodology}
We investigate the internal language representations in multilingual decoder-only LLMs through complementary experimental approaches: logit lens analysis (Section \ref{logit-lens}) and neuron specialization analysis (Section \ref{neuron-specialization}). Each methodology offers unique insights into how models process information across languages. 

\subsection{Models}
We evaluate three models with varying multilingual focus. \textbf{Aya-23-8B } by Cohere AI is an open-source decoder-only model instruction fine-tuned on 23 languages—including ar, zh (simplified \& traditional), en, fr, it, ja, ko, and more—using a two-stage process: pretraining on a balanced multilingual corpus (not public) and multilingual instruction fine-tuning \cite{aryabumi2024aya}. \textbf{Llama 3.1-8B} supports 8 languages (en, fr, de, hi, it, pt, es, th) but was mainly trained on English data (ca. 8\% multilingual tokens) and retains English-centric processing patterns, serving as a baseline for predominantly English-trained models \cite{grattafiori2024llama3herdmodels, wendler2024llamas}. \textbf{Chinese-LLaMA-2-7B} is a Mandarin-adapted LLaMA-2 variant with an expanded tokenizer (+20,000 tokens), pretrained on large Chinese corpora using parameter-efficient fine-tuning (LoRA\cite{hu2021loralowrankadaptationlarge}) and instruction-tuned on millions of Chinese instruction-response pairs, enabling strong Chinese performance at low computational cost \cite{Cui_Efficient_and_Effective_2023, cui2023efficient, hu2021loralowrankadaptationlarge}.

\subsection{Datasets}
In this work, we focus on two primary datasets: the Dumas dataset \cite{dumas2024separating} for logit lens experiments and introduce a new code-mixed dataset that will be publicly released. 

\paragraph{Dumas Dataset}
\label{dumas-ds}
For logit lens experiments, we use the dataset from \citet{dumas2024separating}, which includes word translation and cloze tasks in 13 languages (de, en, es, et, fi, fr, hi, it, ja, ko, nl, ru, zh). It minimizes token overlap between languages while maintaining semantic consistency. Note that model support varies: Aya-23-8B lacks et and fi; Llama 3.1-8B excludes et, fi, ja, ko, nl, ru, and zh; Chinese-LLaMA-2-7B supports only zh and has limited en capabilities, lacking official support for the other 11 languages. Each prompt consists of randomly selected 5-shot word translation examples followed by a final query word. For instance, an English-to-Chinese task may appear as:
\begin{figure}[h!]
\centering
\small 
\fbox{%
  \begin{minipage}{0.8\linewidth}
  \begin{CJK}{UTF8}{gbsn}
  \textit{English: "computer"} → 中文: 电脑 \\
  \textit{English: "ant"} → 中文: 蚂蚁 \\
  \textit{English: "cloud"} → 中文: 云 \\
  \textit{English: "heart"} → 中文: 心脏 \\
  \textit{English: "knife"} → 中文: 刀子 \\
  \textit{English: "book"} → 中文: \_\_
  \end{CJK}
  \end{minipage}%
}
\end{figure}

The task is to predict the correct translation of the final word. Synonyms for the target word are included across all supported languages.

\paragraph{Code-mixed Dataset}
\begin{figure*}[hbt!]
    \centering
    \small
    \includegraphics[width=0.8\linewidth]{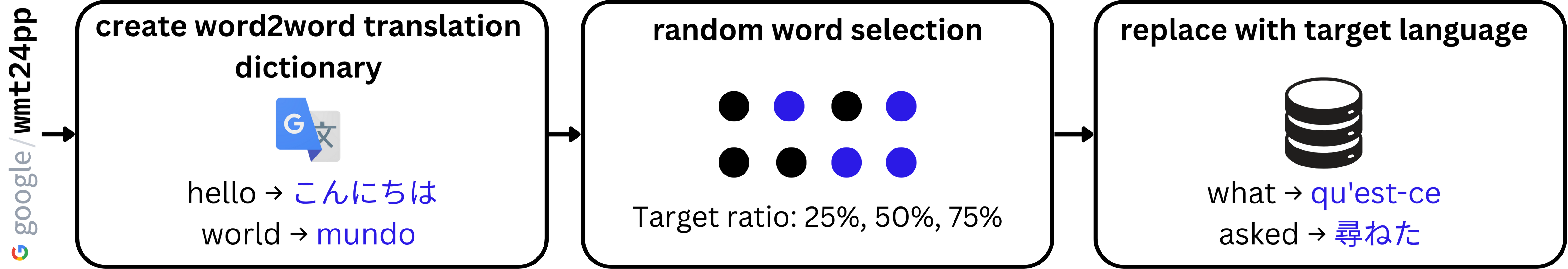}
    \caption{Our code-mixed dataset creation pipeline. Starting with parallel sentences from WMT24++, we create comprehensive bilingual dictionaries using Google Translate for all vocabulary. For each sentence, we randomly select words based on the target mixing ratio (25\%, 50\%, or 75\%) and replace them with their translations in the partner language. For example, from the English source ``The World Bank hopes to spread that message,'' we generate the code-mixed Chinese output ``World bank\begin{CJK}{UTF8}{gbsn}希望传播这一理念\end{CJK}'' (50\%).}
    \label{fig:workflow-codemixed-ds}
\end{figure*}
\label{code-mixed-ds}
To study how models process mixed-language inputs, we construct a code-mixed dataset derived from the WMT24++ parallel corpus \cite{deutsch2025wmt24expandinglanguagecoverage}, containing 998 sentence pairs across 55 languages. We focus on a subsection of 7 languages and take fr and zh as base languages, each mixed with five partner languages (en, es, it, ja, and ko) resulting in ten language pairs. These combinations span a wide typological and script range, including closely related Romance/Indo-European languages (fr/es, fr/it, fr/en), typologically distinct but historically linked pairs (zh/ja, zh/ko), and diverse scripts: Latin (en, fr, es, it), Simplified Chinese (zh), Kanji/Kana (ja), and Hangul (ko).

We generate code-mixed sentences using a three-step rule-based method (Figure \ref{fig:workflow-codemixed-ds}) with controlled mixing ratios of 25\%, 50\%, and 75\%.

We tokenize Latin script using whitespace and Han script with the Jieba library \cite{jieba}. Although this may yield ungrammatical outputs, it ensures consistent mixing ratios critical for controlled experiments. To address limited dictionary coverage in prior work \citep{conneau2017word}, we create comprehensive bilingual dictionaries via Google Translate for all WMT24++ words, ensuring equal vocabulary coverage across language pairs. However, lacking word sense disambiguation, polysemous words are translated identically regardless of context, possibly causing meaning mismatches.

To evaluate translation accuracy, we manually assessed word-level translation quality in code-mixed data, focusing on semantic mistranslations rather than grammatical errors common in code-mixing. From 50\% mixing datasets, we sampled 10 sentences per language pair (246–399 words) and found translation error rates of fr-en 4.76\% , fr-es 4.78\% , zh-en 4.87\% , and zh-es 8.94\% , with higher errors for zh-es due to greater linguistic distance and weaker model performance.

To compare code-mixed and monolingual processing, we include corresponding monolingual datasets from WMT24++ (fr, es, it, ja, and ko) as baselines. 
All code-mixed pairs were evaluated on translation tasks directed from code-mixed input to en (i.e., Chinese-Spanish code-mixed input to en).  
We do not evaluate the reverse direction, as enforcing controlled code-mixing in model-generated outputs is challenging.

To further examine model behavior, we analyze neuron activation patterns (Section~\ref{neuron-specialization}) across code-mixed inputs for Aya-23-8B, LLaMA 3.1-8B, and Chinese-LLaMA-2-7B, testing whether code-mixed processing differs by language pair and model architecture (\textbf{H4}).
%
%

\subsection{Logit Lens}
\label{logit-lens}
Logit lens \citep{nostalgebraist2020} interprets transformer hidden states by projecting intermediate representations into vocabulary space. At each layer $\ell$, the model produces a hidden state $h_\ell \in \mathbb{R}^d$, which is mapped to logits using the unembedding matrix $U \in \mathbb{R}^{|V| \times d}$: $\text{logits}_\ell = U h_\ell$.

These logits approximate the model’s predictions at layer $\ell$. Following \citet{nostalgebraist2020}, we use the residual stream before layer normalization to better align with the final outputs.
Building on prior multilingual analyses \citep{wendler2024llamas, zhong2024englishcentricllmslanguagemultilingual, saji2025romanlensrolelatentromanization}, we apply the logit lens at each layer, extract token probabilities via softmax, and sum over synonyms in 13 languages using the dataset from \citet{dumas2024separating} (see Section \ref{dumas-ds}). To reduce false matches, we apply a 0.1 threshold. This approach allows us to track the emergence of language-specific signals across layers and test \textbf{H1}.

\subsection{Neuron Specialization}
\label{neuron-specialization}

\paragraph{Tan et al.'s Approach}
Following \citet{tan2024neuron}, we identify language-specific neurons via binary ReLU activations in FFNs across WMT24++ and code-mixed data.

For task \(t\) with validation set \(D_t\), each sample \(x_i\) has activation \(\mathbf{a}^t_i\). Summing gives \(\mathbf{a}^t = \sum_{x_i \in D_t} \mathbf{a}^t_i\). Specialized neurons \(S^t_k\) are the top activations satisfying \(\sum_{i \in S^t_k} \mathbf{a}^t(i) \geq k \sum_i \mathbf{a}^t(i)\). Neuron overlap is measured by 
$\mathrm{IoU}(S^i, S^j) = \frac{|S^i \cap S^j|}{|S^i \cup S^j|}$. Using \(k=90\%\) per \citet{tan2024neuron}, we identify neurons covering most activations per language and plot IoU matrices to expose cross-linguistic patterns. Unlike \citet{tan2024neuron}, we exclude neurons shared by all languages to isolate language-specific neurons. This tests \textbf{H2}.




\paragraph{Kojima et al.'s Approach}

\citet{kojima2024multilingual} identified language-specific neurons in multilingual models, concentrated in early and late layers with minimal cross-language sharing.

We extend this to code-mixing neurons in Aya-23-8B's MLP layers. For each code-mixed pair \( l_t \), texts are labeled positive (\(b_i=1\)) or negative (\(b_i=0\)). For neuron \(m\) and text \(x_i = \{w_{i,1}, \ldots, w_{i,T}\}\), activations \(\{z_{m,i,1}, \ldots, z_{m,i,T}\}\) are averaged as \(z_{m,i} = f(z_{m,i,1}, \ldots, z_{m,i,T})\) (excluding padding). We compute Average Precision \(AP_m = AP(z_m, b) \in [0,1]\) to classify neurons into top-\(k\) (high), medium-\(k\) (none), and bottom-\(k\) (negative correlation). Applied to fr and zh code-mixed with en, it, es, jp, ko (10 pairs), this tests \textbf{H3} and \textbf{H4}.




\section{Results and Discussion}

\subsection{Logit Lens Analysis}
To test if balanced multilingual training affects internal processing (\textbf{H1}), we applied logit lens analysis \citep{wendler2024llamas} to Aya-23-8B (balanced), LLaMA 3.1-8B (English-dominant), and Chinese-LLaMA-2-7B (Chinese-specialized).

Using \citet{dumas2024separating}'s dataset, we tracked language-specific token probabilities across layers during translation. From 54 tasks, we computed AUCs for each language probability curve and used Mann-Whitney U tests with Bonferroni correction to compare: (1) model effects -- whether Aya shows more diverse language representations than LLaMA ($p < 0.05/(13 \times 3) = 0.0013$), and (2) task effects -- whether input vs. output languages differ in internal processing ($p < 0.05/(13 \times 3 \times 2) = 0.0006$).

\begin{figure}[t!]
    \centering
    \includegraphics[width=0.8\linewidth]{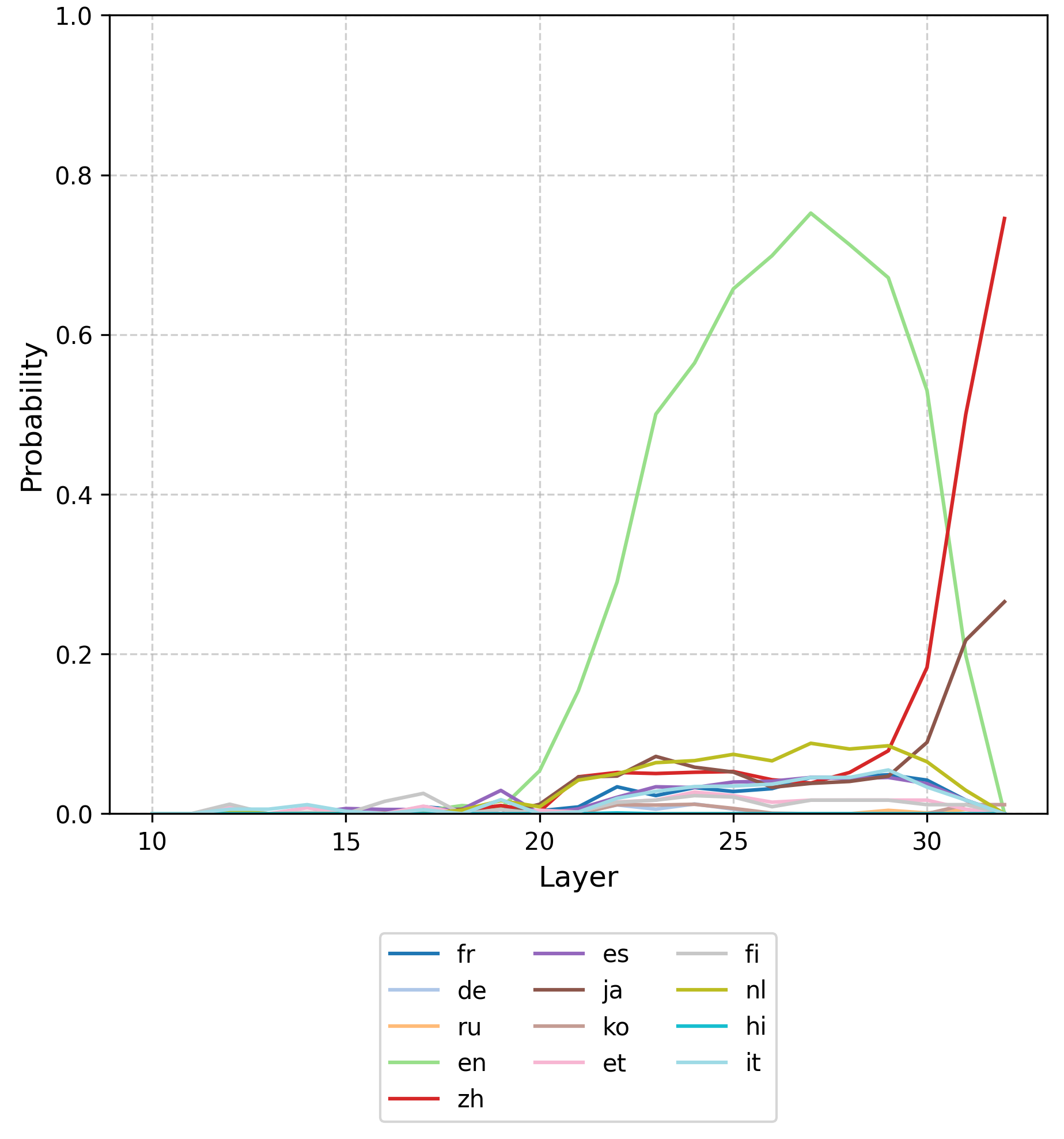}
    \caption{Logit lens language probabilities for English-to-Chinese translation in Aya-23-8B reveal activation of an increased number of languages in mid-to-late layers, with English being dominant.}
    \label{fig:language-probabilities-aya}
\end{figure}

Aya-23-8B demonstrates multilingual processing with cross-linguistic activation. During English-to-Chinese translation (Figure \ref{fig:language-probabilities-aya}), Aya activates multiple languages in intermediate-to-late layers (20–27), including Japanese tokens despite Japanese being neither source nor target. This suggests Aya leverages typological relationships rather than relying solely on English as a pivot.

\begin{figure}[t!]
    \centering
    \includegraphics[width=0.8\linewidth]{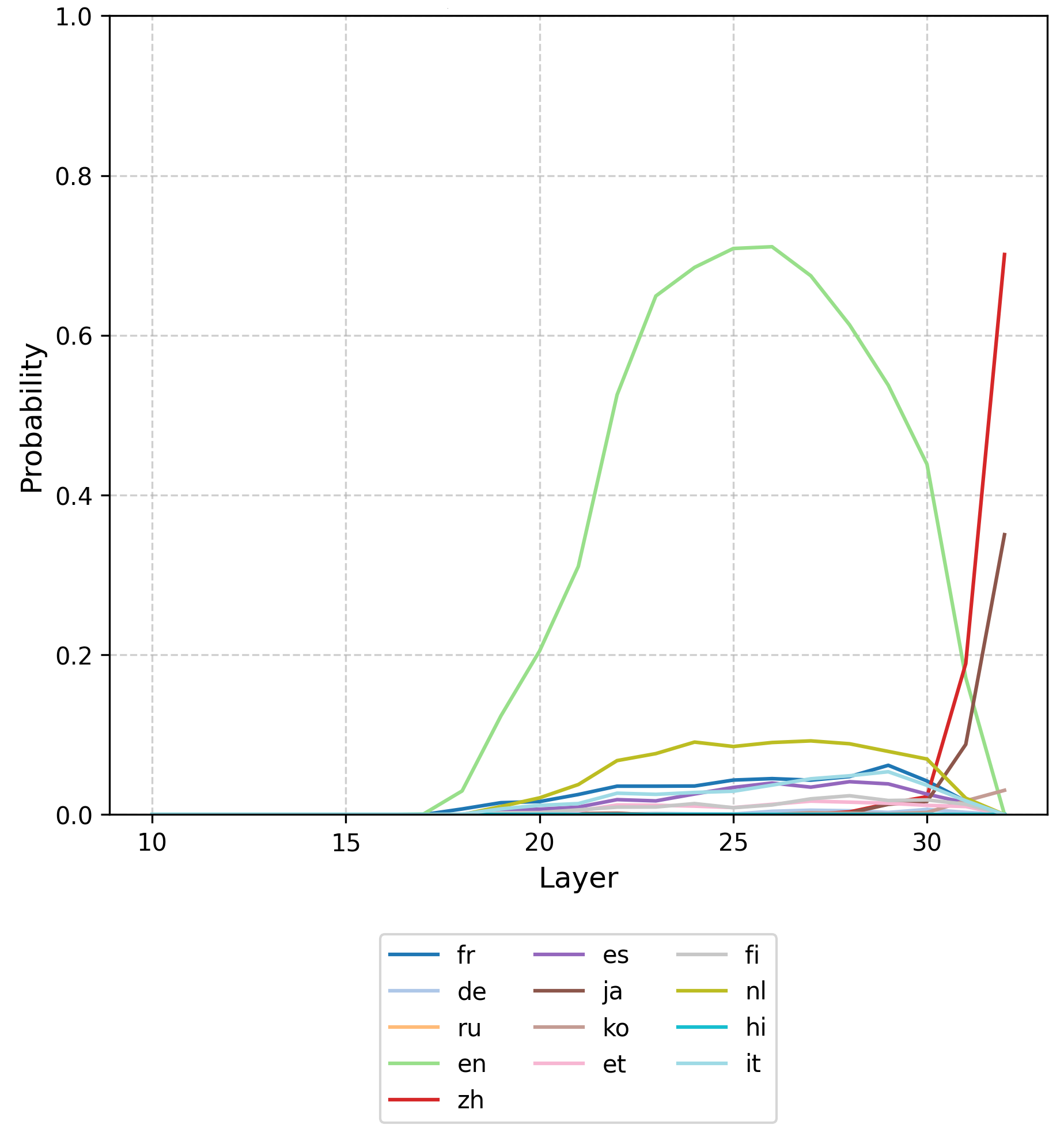}
    \caption{Logit lens language probabilities for English-to-Chinese translation in Llama 3.1-8B show dominant English representations across most layers with few other languages showing significant activation.}
    \label{fig:language-probabilities-llama}
\end{figure}

Llama 3.1-8B follows English-centric processing. In contrast (Figure \ref{fig:language-probabilities-llama}), Llama demonstrates the English-dominated pattern established by \citet{wendler2024llamas}, with English maintaining highest activation across all layers until final output generation. Chinese activates only in final layers, aligning with the ``English-ization'' process \cite{zhao2024large}.

\begin{figure}[h!]
    \centering
    \includegraphics[width=0.8\linewidth]{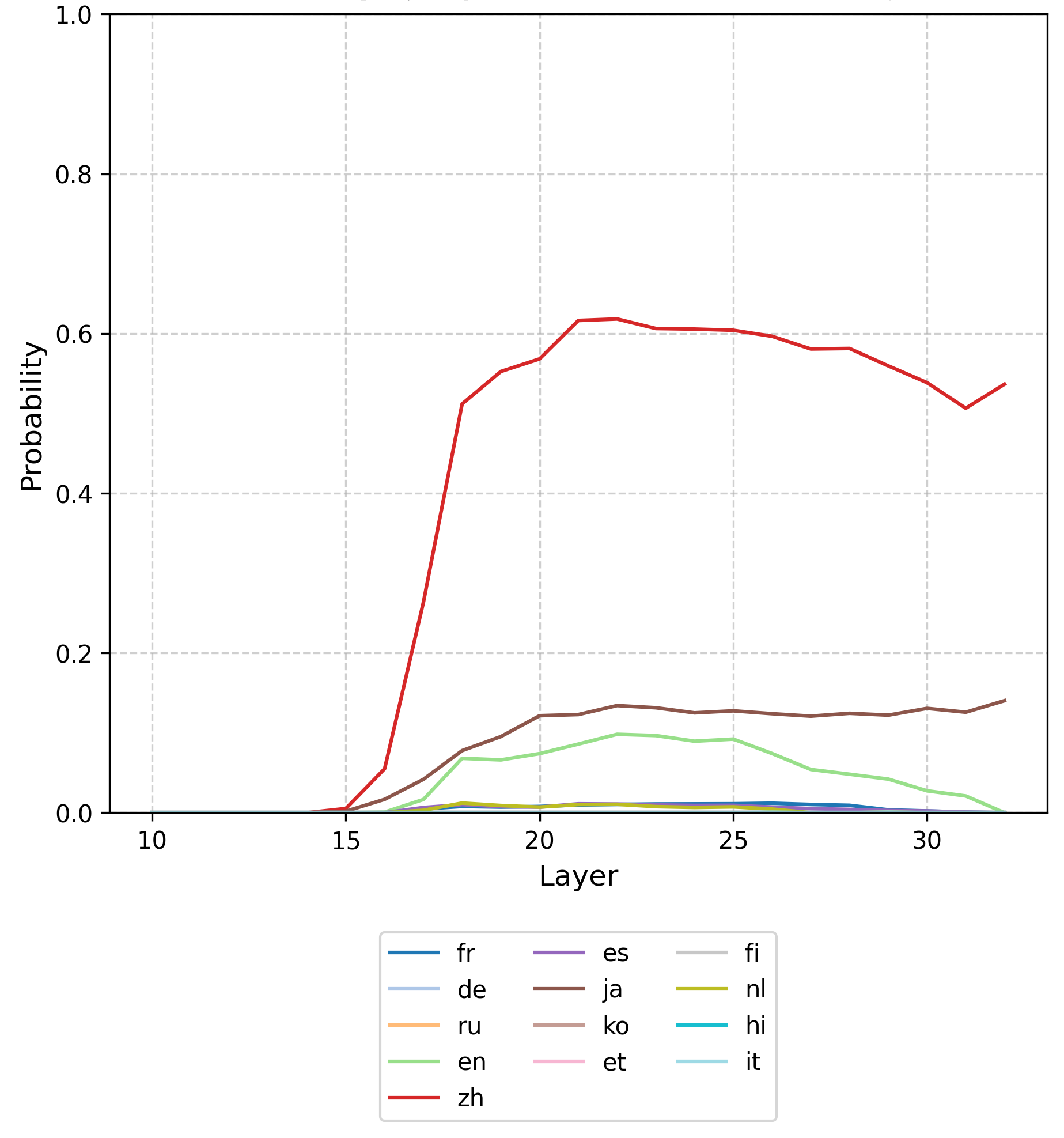}
    \caption{Logit lens language probabilities for English-to-Chinese translation in Chinese-LLaMA-2-7B show strong dominance of Chinese representations across most layers.}
    \label{fig:language-probabilities-chinese}
\end{figure}

Chinese-LLaMA-2-7B exhibits Chinese-dominant processing. This model shows Chinese representations dominating across most layers even for English-to-Chinese translation (Figure \ref{fig:language-probabilities-chinese}), with English activation decreasing in final layers while Japanese remains stable, reflecting its specialized training.

\begin{figure}
    \centering
    \includegraphics[width=1\linewidth]{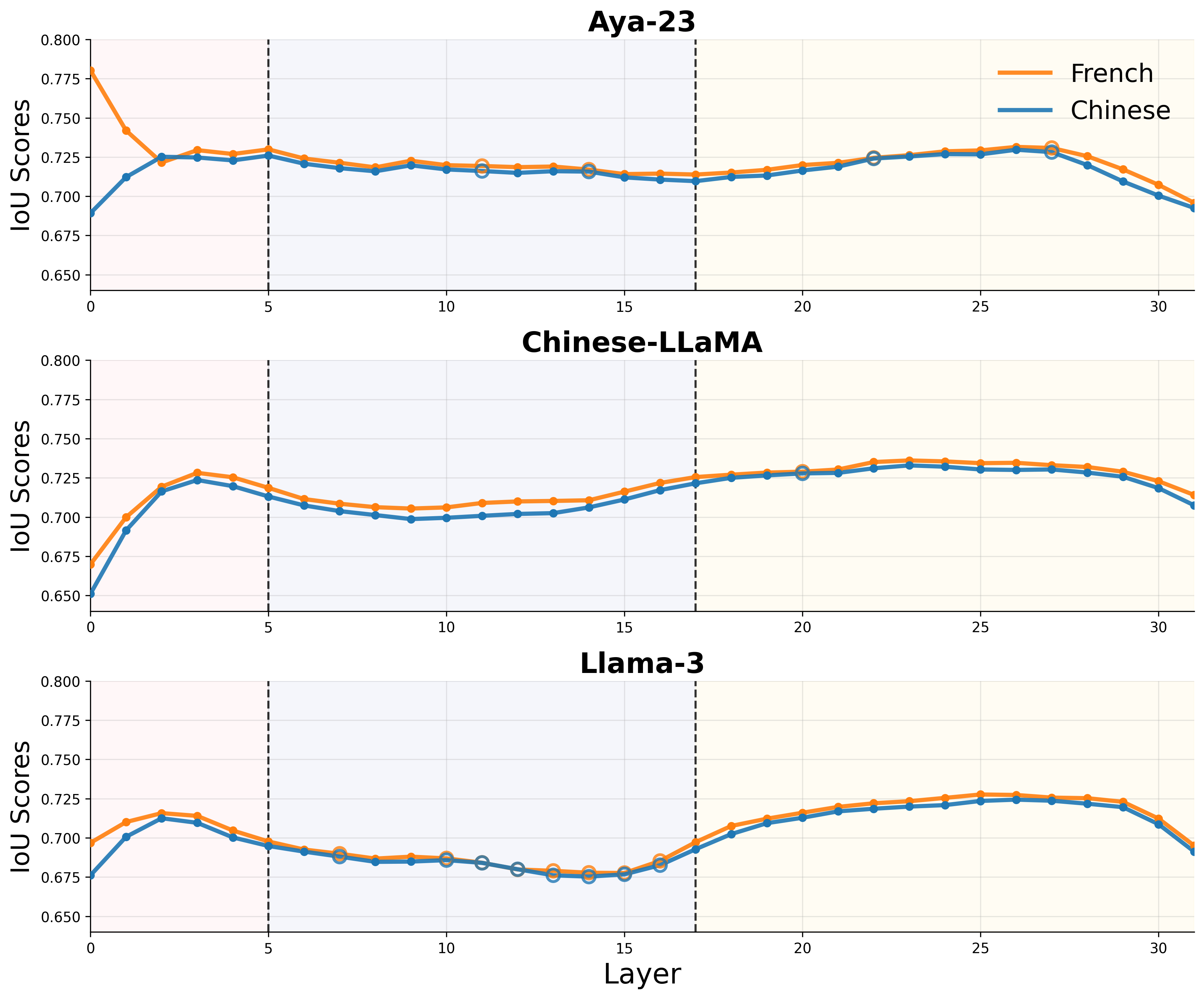}
    \caption{Three-phase neuron clustering patterns across transformer layers. French-based (orange) and Chinese-based (blue) code-mixed language pairs show distinct IoU overlap patterns in Aya-23, Chinese-LLaMA, and Llama-3.1. All models exhibit consistent French processing advantages (86.5\% of layers significant at $p < 0.05$), with Chinese-LLaMA showing the largest effect size ($d=1.466$) despite Chinese specialization. Filled circles indicate statistically significant differences; hollow circles indicate non-significant layers. Phase boundaries at layers 5 and 17 reveal early stabilization, middle-layer divergence, and late convergence patterns.}
    \label{fig:enter-label}
\end{figure}


\begin{figure*}[htbp]
    \centering
    \includegraphics[width=0.8\textwidth]{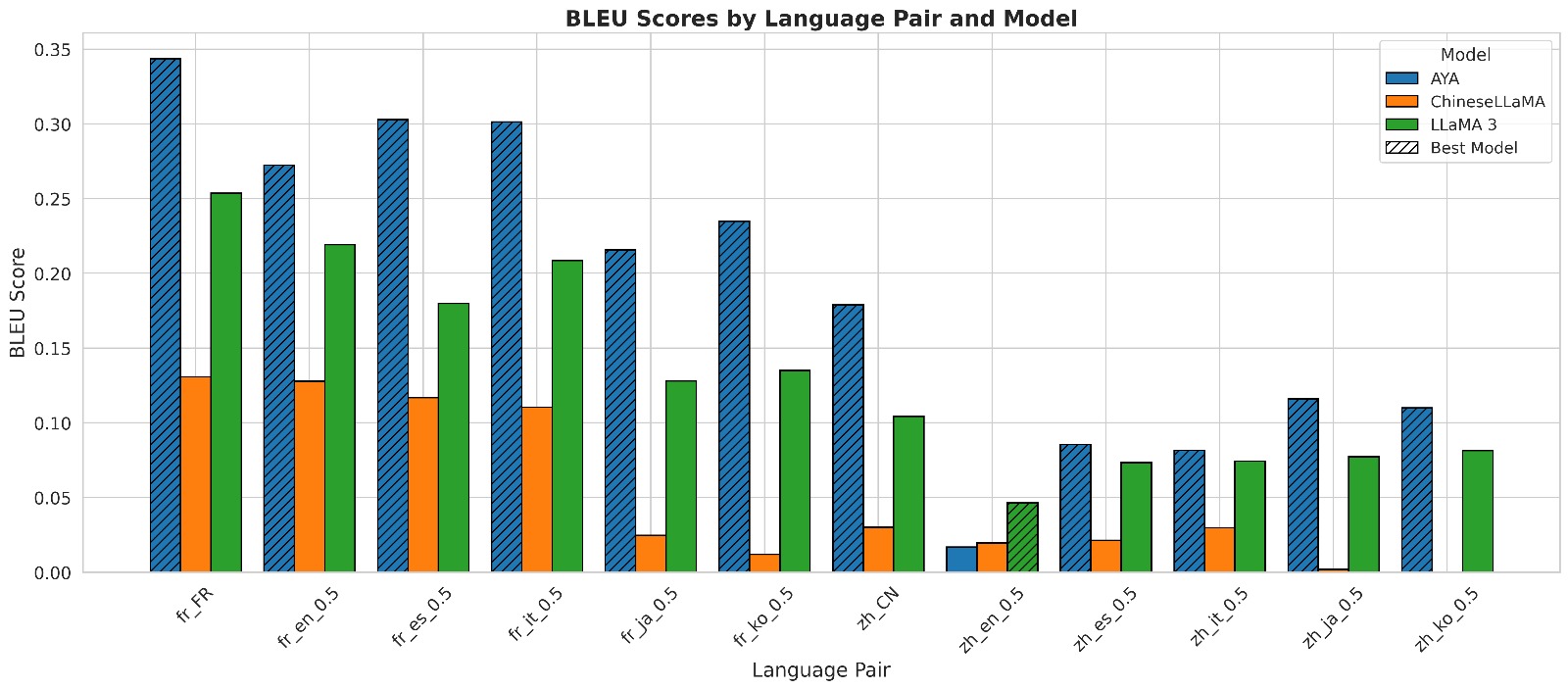}
    \caption{Translation qualities on code-mixed datasets using Aya23-8B, LLaMA 3.1-8B, and Chinese LLaMA, presented in BLEU.}
    \label{fig:translation_quality}
\end{figure*}

Our statistical analysis across all 54 translation tasks provides quantitative support for \textbf{H1}: Aya demonstrates significantly different language activation patterns compared to both Llama (8/13 languages with $p < 0.0013$: de, ru, zh, es, ja, ko, it) and Chinese-LLaMA (8/13 languages including en, zh, es, ja, ko, it). Critically, output languages influence internal representations more strongly than input languages across all models, when analyzing task composition effects, output language presence produces significant changes in 12/13 languages compared to only 7/13 for input languages.


This analysis partially supports our hypothesis that Aya-23 incorporates multiple languages in internal processing, rather than relying solely on English. However, English still shows significantly higher activation probabilities, necessitating careful interpretation of these multilingual patterns. The statistical evidence highlights that both task language and model training paradigm significantly shape internal processing strategies, with task language particularly influencing language-specific activation probabilities.

\subsection{Neuron Specialization Analysis}
\paragraph{Activation \textit{Frequency} Experiments}




Following \citet{tan2024neuron}, we conducted neuron activation frequency experiments to examine how balanced multilingual training influences language-specific processing mechanisms (\textbf{H2}, \textbf{H4}). 

To investigate base-language dependencies systematically, we conducted statistical analysis across all 32 transformer layers using \citet{tan2024neuron}'s methodology. For each layer, we computed IoU overlap values between French-based and Chinese-based code-mixed language pairs, then applied Mann-Whitney U tests to assess whether French-based pairs show significantly higher neuron clustering than Chinese-based pairs. We applied Bonferroni correction ($\alpha = 0.05/32 = 0.0016$) to control for multiple comparisons across layers.

Surprisingly, French-based code-mixed inputs consistently show higher neuron clustering than Chinese-based inputs across all three models (Figure \ref{fig:enter-label}). Aya-23 shows significant French advantage in 28/32 layers (87.5\%) though with a small, average difference of +0.006 IoU; Chinese-LLaMA exhibits the strongest pattern with 31/32 significant layers (96.9\%) and +0.005 average difference; and Llama-3 demonstrates 24/32 significant layers (75\%) with +0.004 average difference. This minor but significant French advantage persists even in Chinese-LLaMA, a model specifically adapted for Chinese processing, challenging our initial hypothesis that base-language training would drive clustering patterns (\textbf{H2}).

Layer-wise analysis reveals three distinct processing phases (Figure \ref{fig:enter-label}). Early layers (0–10) show the strongest French advantages, with Layer 0 having the largest effect (+0.101 in Aya-23). Middle layers (11–21) display increased variability and some non-significant effects, notably in LLaMA-3 where clustering differences are minimal in layers 11–16. Late layers (22–31) consistently show smaller but persistent French advantages across all models. Additionally, a cross-layer trend analysis using Spearman correlation reveals that Chinese-LLaMA shows a moderate decreasing trend ($\rho = -0.408$, $p = 0.021$), while Aya-23 and Llama-3 maintain relatively stable patterns across layers.

Our findings partially contradict \textbf{H2}, as neuron sharing patterns do not align with expected base-language training effects, but strongly support \textbf{H4} code-mixed processing varies systematically with language pair characteristics. The universal French bias (95.8\% of layer comparisons) and distinct script-based clustering patterns demonstrate that typological relationships and orthographic similarities, rather than training data composition, drive neuron activation patterns in multilingual models.

\paragraph{Translation Performance on Code-Mixed Inputs}
Figure \ref{fig:translation_quality} presents BLEU scores for all three models on monolingual and code-mixed datasets. Aya-23-8B consistently outperforms the others, with a clear advantage on fr-based code-mixed inputs. All models show better performance on Latin-script pairs (fr-en, fr-es, fr-it) than on cross-script ones (fr-ja, fr-ko). For zh code-mixing, Aya-23-8B and Llama 3.1-8B perform better on zh-ja and zh-ko than on zh-en, zh-fr, and zh-it, suggesting that shared vocabulary and typological features help transfer despite script differences. In contrast, Chinese-LLaMA-2-7B performs poorly across all code-mixed inputs, regardless of typological similarity.

Performance generally degrades as code-mixing rate increases across all models, likely reflecting limitations of our rule-based word-to-word translation approach. However, Aya-23-8B shows greater resilience to this degradation, supporting our finding that balanced multilingual training improves robustness to code-mixing. 

\begin{figure}[h!]
    \centering
    \includegraphics[width=0.7\linewidth]{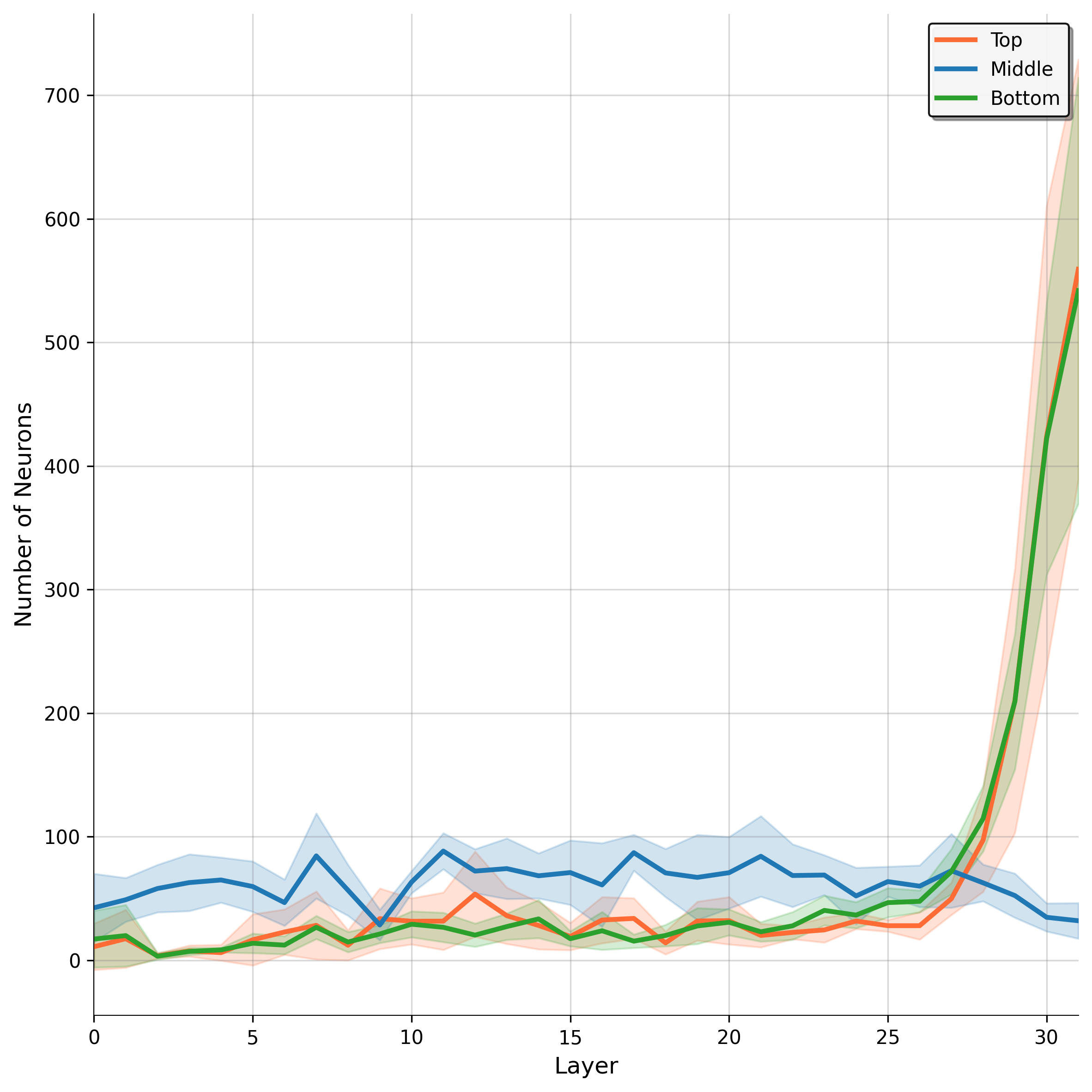}
    \caption{Layer-wise distribution of $k=1000$ language-specific neurons in Aya-23-8B for code-mixed processing across all CM language pairs in Aya-23-8B.}
    \label{fig:distribution}
\end{figure}

\paragraph{Activation \textit{Strength} Experiments}
To address H4, we followed \citet{kojima2024multilingual}'s methodology by processing both monolingual and code-mixed texts and capturing neuron activations at the MLP layers. Our findings for Aya reveal an interesting divergence from previous work on decoder-only model.
While \citet{kojima2024multilingual} found language-specific neurons (both top-k and bottom-k) concentrated in first and last layers of other decoder-only models, Aya-23-8B exhibits a different pattern when processing code-mixed input: top-k language-specific neurons appear predominantly in final layers (27-31), with a pronounced spike in layer 31 across all language pairs (see Figure \ref{fig:distribution}). This pattern confirms our hypothesis \textbf{H3}.

\begin{figure}[h!]
    \centering
    \includegraphics[width=0.8\linewidth]{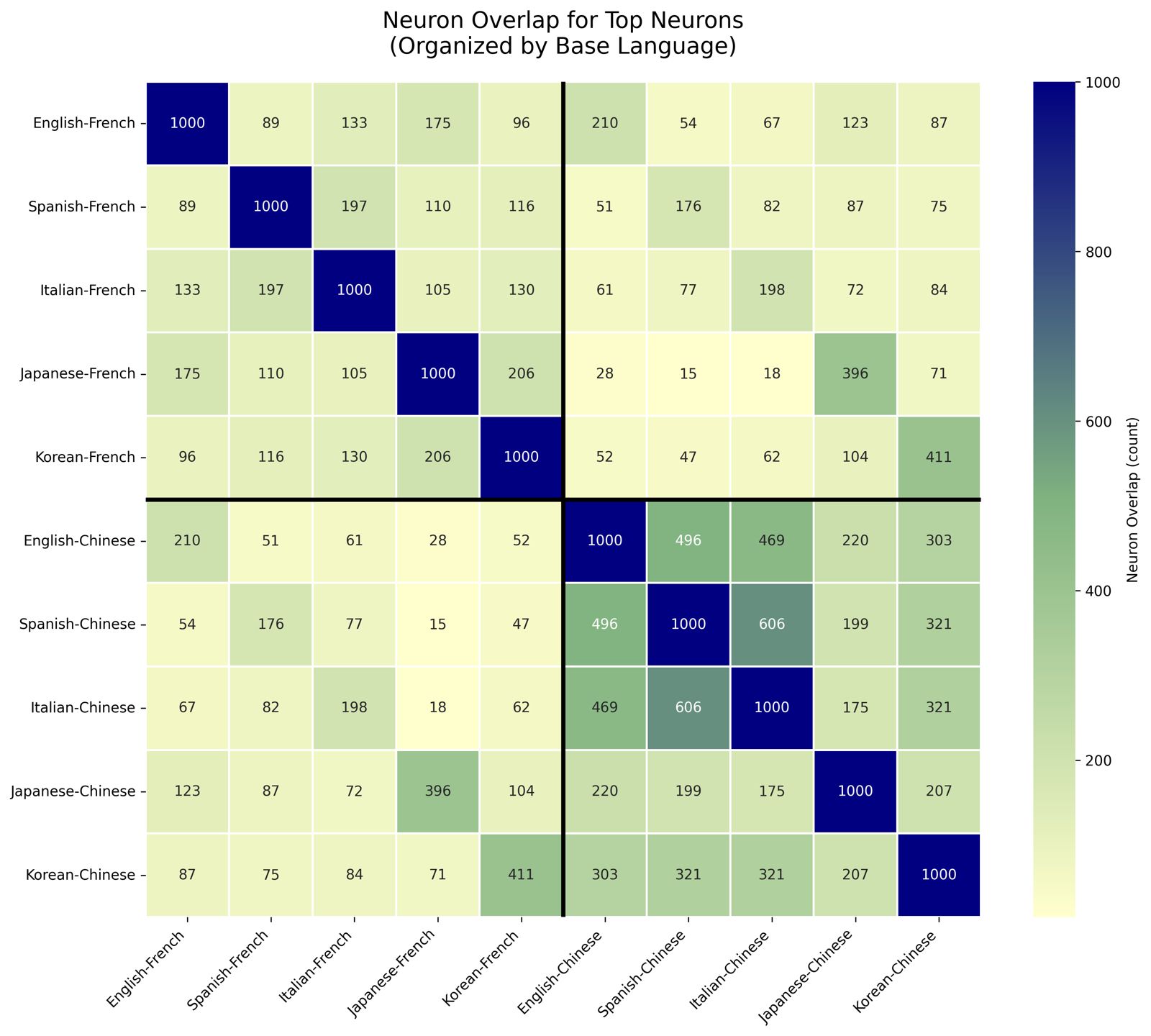}
    \caption{The number of overlapping language-specific neurons between code-mixing language pairs in Aya-23-8B.}
    \label{fig:overlap-top-k}
\end{figure}

This pattern only partially aligns with \citet{tang2024language}, who observed a skewed ``U''-shaped distribution, with language processing concentrated in both early and late layers. In contrast, it supports the findings of \citet{mondal2025language}, who reported that language-specific neurons in modern LLMs are primarily concentrated in later layers. Our results suggest that Aya-23-8B’s balanced multilingual training may promote a shift toward language-specific processing concentrated at the generation stage, diverging from the more distributed patterns seen in predominantly monolingual models.

This pattern remains consistent across all language pairs. Bottom-k neurons (``anti-correlated'' neurons) similarly concentrate in final layers, while medium-k neurons distribute more evenly across early (0-5) and middle (10-20) layers. 

This distinctive concentration pattern may stem from Aya’s explicitly balanced multilingual training, resulting in an internal structure different from the predominantly monolingual models studied by \citet{kojima2024multilingual}. The pronounced spike of language-specific neurons in the final layer likely reflects Aya-23-8B’s processing strategy for code-mixed inputs: earlier layers handle distributed multilingual representations for understanding, while the concentration at the generation stage resolves which language to output in mixed-language contexts.


Our analysis of neuron overlap shows that the base language influences neuron sharing more than the secondary mixed-in language. Chinese-based pairs consistently exhibit higher neuron overlap (17.5\%–60.6\%) compared to French-based pairs, regardless of the secondary language (Figure \ref{fig:overlap-top-k}). This indicates that the foundational language’s structural properties strongly shape neural organization. This pattern holds on average (French-based pairs: 138.25 neurons; Chinese-based pairs: 331.7 neurons; cross-base pairs: 82.65 neurons)\footnote{Notable exceptions exist where typological similarity overrides base language effects, such as fr-ja with zh-ja (396 neurons) and ko-fr with ko-zh (411 neurons), likely reflecting historical Japanese-Chinese and Korean-Chinese linguistic contact}. 

Cross-script connections also appear, with ja-zh and ko-zh pairs showing moderate neuron overlap (20.7\% and 41.1\%, respectively), likely due to shared vocabulary and writing systems from historical contact. Within the fr-based group, neuron sharing varies: it-fr pairs have the highest overlap (19.7\%), followed by en-fr (17.5\%) and es-fr (8.9\%), suggesting that typological similarity within the Romance family shapes neural processing patterns.

\section{Related Work}
\paragraph{Pivot Languages in Multilingual LLMs}
Training data composition fundamentally shapes multilingual processing patterns. Llama models, heavily trained on English (89\% in Llama-2 \cite{touvron2023llama}), use English as a ``pivot language'' in multilingual tasks -- translating French to English before Chinese, reducing quality \cite{wendler2024llamas}. This English bias extends beyond translation, with models defaulting to English in intermediate layers for reasoning \cite{zhao2024large, zhong2024beyond}. The Multilingual Workflow (MWork) hypothesis \cite{zhao2024large} formalizes this as: convert inputs to English for reasoning, integrate multilingual knowledge, then generate target output.

However, English-centric processing varies with architecture and training. Language-specific models like Swallow (Japanese-adapted Llama-2) and LLM-jp default to their dominant training language rather than English \cite{zhong2024beyond}. \citet{schut2025multilingual} found Aya-23 activated English ca. 50\% versus ca. 70\% in Gemma-2-27B, suggesting balanced training reduces English dominance. Similarly, \citet{lindsey2025biology} identified language-agnostic conceptual representations in Claude 3.5 Haiku, indicating some models develop universal processing spaces beyond pivot strategies.

\paragraph{Language-Specific Neurons}
Language-specific neurons in decoder-only models cluster distinctly with minimal cross-language sharing. \citet{kojima2024multilingual} and \citet{tang2024language} found these neurons concentrate in top and bottom layers of LLaMA-2, BLOOM, and Mistral, comprising only 1\% of parameters. However, \citet{mondal2025language} observed newer models (Mistral Nemo, Llama 3.1) concentrate language-specific neurons primarily in later layers, indicating architectural evolution. Training data biases models toward English, degrading performance with increasing linguistic distance \cite{zhong2024beyond, wendler2024llamas}, though positive cross-lingual transfer remains possible.

Recent work reveals dynamic language-specific processing. \citet{tan2024neuron} found feed-forward neurons in encoder-decoder models activate in language-specific patterns, with overlaps reflecting linguistic proximity. \citet{deng2025unveiling} demonstrated that models dynamically shift activations based on context -- Spanish prefixes amplify Spanish-specific features while suppressing others -- suggesting sophisticated contextual language processing beyond fixed neuron assignments.

\paragraph{Code-Mixing and Script-Based Processing}
Code-mixing (CM) research reveals systematic biases in multilingual processing. \citet{wang2025language} showed reasoning language models activate Latin and Han scripts even when processing Arabic, Hindi, or Japanese, with performance gains up to 110\% when constraining reasoning to preferred scripts. This suggests script-based processing preferences shaped by training data composition.

CM poses significant challenges for multilingual LLMs, particularly for low-resource languages. \citet{gupta2024code} found GPT models perform worse on English-Gujarati CM compared to English-French, reflecting training data imbalances toward high-resource monolingual corpora \cite{gundapu2020word}. \citet{yang-etal-2020-csp} demonstrated CM-specific pre-training improves translation performance, indicating models can learn to handle language transitions within utterances.

Our study addresses the underexplored gap between predominantly English-trained models (Llama) and balanced multilingual models (Aya-23), investigating whether reduced English reliance corresponds to distinct internal architectures through comprehensive neuron-level analysis across languages and code-mixed contexts.

\section{Conclusion}
Our investigation reveals that balanced multilingual training fundamentally alters how decoder-only LLMs process language internally. Through logit lens analysis, we show that Aya-23-8B employs distinct multilingual processing strategies, activating typologically related languages (e.g., Japanese during Chinese translation) and exhibiting significantly different activation patterns compared to English-centric models across 8/13 languages. We find that output languages influence internal representations more strongly than input languages.

Our neuron specialization analysis reveals that Aya-23-8B concentrates language-specific neurons predominantly in final layers (27-31) rather than distributing them across early and late layers as found in previous studies of decoder-only models \cite{kojima2024multilingual, tang2024language}. This architectural difference suggests that balanced multilingual training creates models that maintain language-agnostic processing through most layers, with language-specific differentiation emerging primarily at generation time.

Code-mixed processing reveals systematic patterns driven by base language characteristics and script similarity. Base languages drive neuron sharing more strongly than mixed-in languages, with French-based code-mixed inputs maintaining consistent neuron overlap regardless of mixing rate, while Chinese-based inputs show proportional degradation. Translation performance demonstrates clear advantages for same-script language pairs, though Chinese-Japanese and Chinese-Korean pairs benefit from shared historical vocabulary despite script differences.

\section*{Limitations}
Our study has several important limitations that should be considered when interpreting the results. In our logit lens experiments, despite efforts to minimize token overlap between languages, some overlap likely remains between Japanese and Chinese, as well as between French and English. Though this overlap is minimal, it could potentially affect our analysis of language-specific activation patterns.

For our implementation of Tan et al.'s neuron specialization analysis, we observed relatively weak sharing patterns in our heatmap visualizations, which may limit the strength of our conclusions regarding language-specific processing. Additionally, our methodology requires converting continuous neuron activations to binary values, resulting in information loss that potentially obscures subtle activation patterns across languages.

The quality of our code-mixed dataset represents another limitation. We created code-mixed inputs using rule-based word-to-word translation, which fails to account for grammatical structure and often produces unnatural sentences that may not accurately represent authentic code-switching behavior.

Our findings related to Kojima et al.'s approach reveal an interesting discrepancy. While Kojima found language-specific neurons concentrated in both first and last layers of decoder-only models, our analysis of Aya-23-8B on code-mixed input shows language-specific neurons predominantly in the final layers (27-31), with a pronounced spike in layer 31. This difference likely stems from the fact that what we measure is better characterized as ``code-mixing neurons'' rather than pure language neurons, since our classification task distinguished between code-mixed and non-code-mixed inputs. Our results suggest that code-mixing neurons differ from language neurons, with similar processing in early layers but significant divergence in later layers. This means that regarding hypothesis H4, we can only conclude that code-mixed inputs are processed differently in the very late layers of the model. Similarly, our findings from the Tan et al. experiment indicate that processing varies by language pair across all layers, but without establishing a clear pattern related to language families or script types that would fully support hypothesis H3.

Our analysis is limited to three models (Aya-23-8B, Llama 3.1-8B, and Chinese-LLaMA-2-7B) and may not generalize to other multilingual model architectures or sizes. We also focus primarily on high-resource languages, with limited investigation into how low-resource languages are processed internally.

\section*{Ethics Statement}
We identify no ethical concerns directly related to this research. All models and datasets used in this study are employed in accordance with their respective license terms, including the custom use license for Llama 3.1-8B, the Apache 2.0 license for Aya-23-8B, and the research-permitted use of Chinese-LLaMA-2-7B. The Dumas dataset and WMT24++ corpus are used under their standard research licenses. Our code-mixed dataset, created through rule-based translation, contains no sensitive personal information and will be made publicly available to support reproducible research. The neuron-level analysis conducted in this work focuses purely on model internals without generating potentially harmful content or reinforcing linguistic biases.

\bibliography{anthology,custom}
\bibliographystyle{acl_natbib}

\appendix

\newpage

\end{document}